\pdfoutput=1

\documentclass[11pt]{article}

\usepackage[]{acl}

\usepackage{times}
\usepackage{latexsym}

\usepackage[T1]{fontenc}

\usepackage[utf8]{inputenc}

\usepackage{microtype}
\usepackage{amsmath}

\usepackage{multirow}
\usepackage{multicol}
\usepackage{makecell}
\usepackage{graphicx} 
\usepackage{amssymb}
\usepackage{algorithm}
\usepackage{algpseudocode}
\usepackage{ulem}
\usepackage{booktabs}
\usepackage{CJKutf8}
\usepackage{subfigure}

\usepackage{arydshln}
\usepackage{enumitem}

\title{Bi-SimCut: A Simple Strategy for Boosting Neural Machine Translation}

\author{Pengzhi Gao, Zhongjun He, Hua Wu, and Haifeng Wang \\
Baidu Inc. No. 10, Shangdi 10th Street, Beijing, 100085, China \\
\texttt{\{gaopengzhi,hezhongjun,wu\_hua,wanghaifeng\}@baidu.com} 
}

\begin{document}
\maketitle

\begin{abstract}
We introduce Bi-SimCut: a simple but effective training strategy to boost neural machine translation (NMT) performance. It consists of two procedures: bidirectional pretraining and unidirectional finetuning. Both procedures utilize SimCut, a simple regularization method that forces the consistency between the output distributions of the original and the cutoff sentence pairs. Without leveraging extra dataset via back-translation or integrating large-scale pretrained model, Bi-SimCut achieves strong translation performance across five translation benchmarks (data sizes range from 160K to 20.2M): BLEU scores of $31.16$ for $\texttt{en}\rightarrow\texttt{de}$ and $38.37$ for $\texttt{de}\rightarrow\texttt{en}$ on the IWSLT14 dataset, $30.78$ for $\texttt{en}\rightarrow\texttt{de}$ and $35.15$ for $\texttt{de}\rightarrow\texttt{en}$ on the WMT14 dataset, and $27.17$ for $\texttt{zh}\rightarrow\texttt{en}$ on the WMT17 dataset\footnote{Source code: https://github.com/gpengzhi/Bi-SimCut}. SimCut is not a new method, but a version of Cutoff \cite{shen2020simple} simplified and adapted for NMT, and it could be considered as a perturbation-based method. Given the universality and simplicity of SimCut and Bi-SimCut, we believe they can serve as strong baselines for future NMT research.
\end{abstract}

\section{Introduction}

The state of the art in machine translation has been dramatically improved over the past decade thanks to the neural machine translation (NMT) \cite{wu2016google}, 
and Transformer-based models \cite{vaswani2017attention} often deliver state-of-the-art (SOTA) translation performance with large-scale corpora \cite{ott2018scaling}.
Along with the development in the NMT field, consistency training \cite{bachman2014learning} has been widely adopted and shown great promise to improve NMT performance. It simply regularizes the NMT model predictions to be invariant to either small perturbations applied to the inputs \cite{sano2019effective,shen2020simple} and hidden states \cite{chen2021manifold} or the model randomness and variance existed in the training procedure \cite{liang2021r}. 

Specifically, \citet{shen2020simple} introduce a set of cutoff data augmentation methods and utilize Jensen-Shannon (JS) divergence loss to force the consistency between the output distributions of the original and the cutoff augmented samples in the training procedure. Despite its impressive performance, finding the proper values for the four additional hyper-parameters introduced in cutoff augmentation seems to be tedious and time-consuming if there are limited resources available, which hinders its practical value in the NMT field. 

In this paper, our main goal is to provide a simple, easy-to-reproduce, but tough-to-beat strategy for training NMT models. Inspired by cutoff augmentation \cite{shen2020simple} and virtual adversarial regularization \cite{sano2019effective} for NMT, we firstly introduce a simple yet effective regularization method named SimCut. Technically, SimCut is not a new method and can be viewed as a simplified version of Token Cutoff proposed in \citet{shen2020simple}. We show that bidirectional backpropagation in Kullback-Leibler (KL) regularization plays a key role in improving NMT performance. We also regard SimCut as a perturbation-based method and discuss its robustness to the noisy inputs. At last, motivated by bidirectional training \cite{ding2021improving} in NMT, we present Bi-SimCut, a two-stage training strategy consisting of bidirectional pretraining and unidirectional finetuning equipped with SimCut regularization.

The contributions of this paper can be summarized as follows:
\begin{itemize}
\item We propose a simple but effective regularization method, SimCut, for improving the generalization of NMT models. SimCut could be regarded as a perturbation-based method and serves as a strong baseline for the approaches of robustness. We also show the compatibility of SimCut with the pretrained language models such as mBART \cite{liu2020multilingual}.

\item We propose Bi-SimCut, a training strategy for NMT that consists of bidirectional pretraining and unidirectional finetuning with SimCut regularization.

\item Our experimental results show that NMT training with Bi-SimCut achieves significant improvements over the Transformer model on five translation benchmarks (data sizes range from 160K to 20.2M), and outperforms the current SOTA method BiBERT \cite{xu-etal-2021-bert} on several benchmarks.
\end{itemize}

\section{Background}

\subsection{Neural Machine Translation}

The NMT model refers to a neural network with an encoder-decoder architecture, which receives a sentence as input and returns a corresponding translated sentence as output. Assume $\mathbf{x} = x_1, ..., x_I$ and $\mathbf{y} = y_1, ..., y_J$ that correspond to the source and target sentences with lengths $I$ and $J$ respectively. Note that $y_J$ denotes the special end-of-sentence symbol $\langle eos \rangle$. The encoder first maps a source sentence $\mathbf{x}$ into a sequence of word embeddings $e(\mathbf{x}) = e(x_1), ..., e(x_I)$, where $e(\mathbf{x}) \in \mathbb{R}^{d \times I}$, and $d$ is the embedding dimension. The word embeddings are then encoded to the corresponding hidden representations $\mathbf{h}$. Similarly, the decoder maps a shifted copy of the target sentence $\mathbf{y}$, i.e., $\langle bos \rangle, y_1, ..., y_{J-1}$, into a sequence of word embeddings $e(\mathbf{y}) = e(\langle bos \rangle), e(y_1), ..., e(y_{J-1})$, where $\langle bos \rangle$ denotes a special beginning-of-sentence symbol, and $e(\mathbf{y}) \in \mathbb{R}^{d \times J}$. The decoder then acts as a conditional language model that operates on the word embeddings $e(\mathbf{y})$ and the hidden representations $\mathbf{h}$ learned by the encoder.

Given a parallel corpus $\mathcal{S} = \{\mathbf{x}^i, \mathbf{y}^i\}_{i=1}^{|\mathcal{S}|}$, the standard training objective is to minimize the empirical risk:
\begin{equation}
\mathcal{L}_{ce}(\theta) =  \mathop{\mathbb{E}}\limits_{(\mathbf{x}, \mathbf{y}) \in \mathcal{S}} [\ell(f(\mathbf{x}, \mathbf{y}; \theta), \ddot{\mathbf{y}})],
\end{equation}
where $\ell$ denotes the cross-entropy loss, $\theta$ is a set of model parameters, $f(\mathbf{x}, \mathbf{y}; \theta)$ is a sequence of probability predictions, i.e., 
\begin{equation}
f_j(\mathbf{x}, \mathbf{y}; \theta) = P(y|\mathbf{x}, \mathbf{y}_{<j}; \theta),
\end{equation}
and $\ddot{\mathbf{y}}$ is a sequence of one-hot label vectors for $\mathbf{y}$. 

\subsection{Cutoff Augmentation}

\citet{shen2020simple} introduce a set of cutoff methods which augments the training by creating the partial views of the original sentence pairs and propose Token Cutoff for the machine translation task. Given a sentence pair $( \mathbf{x}, \mathbf{y})$, $N$ cutoff samples $\{ \mathbf{x}^i_{\rm cut}, \mathbf{y}^i_{\rm cut} \}_{i=1}^N$ are constructed by randomly setting the word embeddings of $x_1, ..., x_I$ and $y_1, ..., y_J$ to be zero with a cutoff probability $p_{\rm cut}$. For each sentence pair, the training objective of Token Cutoff is then defined as:
\begin{equation}\label{token_cutoff}
\mathcal{L}_{tokcut}(\theta) = \mathcal{L}_{ce}(\theta) + \alpha \mathcal{L}_{cut}(\theta) + \beta \mathcal{L}_{kl}(\theta),
\end{equation}
where
\begin{equation}
\mathcal{L}_{ce}(\theta) = \ell(f(\mathbf{x}, \mathbf{y}; \theta), \ddot{\mathbf{y}}),
\end{equation}
\begin{equation}
\mathcal{L}_{cut}(\theta) = \frac{1}{N}\sum_{i=1}^N\ell(f(\mathbf{x}^i_{\rm cut}, \mathbf{y}^i_{\rm cut}; \theta), \ddot{\mathbf{y}}),
\end{equation}
\begin{align}
\mathcal{L}_{kl}(\theta) = & \frac{1}{N+1} \{ \sum_{i=1}^N \text{KL}(f(\mathbf{x}^i_{\rm cut}, \mathbf{y}^i_{\rm cut}; \theta) \| p_{\rm avg}) \nonumber \\
& + \text{KL}(f(\mathbf{x}, \mathbf{y}; \theta) \| p_{\rm avg}) \},
\end{align}
\begin{align}
p_{\rm avg} = & \frac{1}{N+1} \{ \sum_{i=1}^N f(\mathbf{x}^i_{\rm cut}, \mathbf{y}^i_{\rm cut}; \theta) \nonumber \\
& + f(\mathbf{x}, \mathbf{y}; \theta) \},
\end{align}
in which $\text{KL}(\cdot \| \cdot)$ denotes the Kullback-Leibler (KL) divergence of two distributions, and $\alpha$ and $\beta$ are the scalar hyper-parameters that balance $\mathcal{L}_{ce}(\theta)$, $\mathcal{L}_{cut}(\theta)$ and $\mathcal{L}_{kl}(\theta)$.

\section{Datasets and Baseline Settings}\label{data_baseline}

In this section, we describe the datasets used in experiments as well as the model configurations. For fair comparisons, we keep our experimental settings consistent with previous works.

\paragraph{Datasets}

\begin{table}[h]
\centering
\begin{tabular}{c|c|c|c} 
 & IWSLT & \multicolumn{2}{|c}{WMT} \\
\hline
\hline
& \texttt{en}$\leftrightarrow$\texttt{de} & \texttt{en}$\leftrightarrow$\texttt{de} & \texttt{zh}$\rightarrow$\texttt{en} \\
\hline
train & 160239 & 4468840 & 20184941 \\ 
valid & 7283 & 6003 & 2002 \\ 
test & 6750 & 3003 & 2001 \\ 
\hline
\end{tabular}
\caption{Number of sentence pairs used in our machine translation experiments.}
\label{statics}
\end{table}

We initially consider a low-resource (IWSLT14 \texttt{en}$\leftrightarrow$\texttt{de}) scenario and then show further experiments in standard (WMT14 \texttt{en}$\leftrightarrow$\texttt{de}) and high (WMT17 \texttt{zh}$\rightarrow$\texttt{en}) resource scenarios in Sections \ref{standard-resource} and \ref{high-resource}. The detailed information of the datasets are summarized in Table~\ref{statics}. We here conduct experiments on the IWSLT14 English-German dataset\footnote{https://github.com/pytorch/fairseq/blob/main/examples/ translation/prepare-iwslt14.sh}, which has 160K parallel bilingual sentence pairs. Following the common practice, we lowercase all words in the dataset. We build a shared dictionary with 10K byte-pair-encoding (BPE) \cite{sennrich2015neural} types.

\paragraph{Settings}

We implement our approach on top of the Transformer \cite{vaswani2017attention}. We apply a Transformer with 6 encoder and decoder layers, 4 attention heads, embedding size 512, and FFN layer dimension 1024. We apply cross-entropy loss with label smoothing rate $0.1$ and set max tokens per batch to be $4096$. 
We use Adam optimizer with Beta $(0.9, 0.98)$, $4000$ warmup updates, and inverse square root learning rate scheduler with initial learning rates $5e^{-4}$. We use dropout rate $0.3$ and beam search decoding with beam size $5$ and length penalty $1.0$. We apply the same training configurations in both pretraining and finetuning stages which will be discussed in the following sections. We use $\texttt{multi-bleu.pl}$\footnote{https://github.com/moses-smt/mosesdecoder/blob/ master/scripts/generic/multi-bleu.perl} for BLEU \cite{papineni2002bleu} evaluation. We train all models until convergence on a single NVIDIA Tesla V100 GPU. All reported BLEU scores are from a single model. For all the experiments below, we select the saved model state with the best validation performance.

\section{Bi-SimCut}

In this section, we formally propose Bidirectional Pretrain and Unidirectional Finetune with Simple Cutoff Regularization (Bi-SimCut), a simple but effective training strategy that can greatly enhance the generalization of the NMT model. Bi-SimCut consists of a simple cutoff regularization and a two-phase pretraining and finetuning strategy. We introduce the details of each part below.

\subsection{SimCut: A Simple Cutoff Regularization for NMT}

Despite the impressive performance reported in \citet{shen2020simple}, finding the proper hyper-parameters $(p_{\rm cut}, \alpha, \beta, N)$ in Token Cutoff seems to be tedious and time-consuming if there are limited resources available, which hinders its practical value in the NMT community. To reduce the burden in hyper-parameter searching, we propose SimCut, a simple regularization method that forces the consistency between the output distributions of the original sentence pairs and the cutoff samples. 

Our problem formulation is motivated by Virtual Adversarial Training (VAT), where \citet{sano2019effective} introduces a KL-based adversarial regularization that forces the output distribution of the samples with adversarial perturbations $\boldsymbol \delta_{\mathbf{x}} \in \mathbb{R}^{d \times I}$ and $\boldsymbol \delta_{\mathbf{y}} \in \mathbb{R}^{d \times J}$ to be consistent with that of the original samples:
\begin{equation*}
\text{KL} (f(e(\mathbf{x}), e(\mathbf{y});\theta) \| f(e(\mathbf{x}) + \boldsymbol \delta_{\mathbf{x}}, e(\mathbf{y}) + \boldsymbol \delta_{\mathbf{y}}; \theta)).
\end{equation*}
Instead of generating perturbed samples by gradient-based adversarial methods, for each sentence pair $( \mathbf{x}, \mathbf{y} )$, we only generate one cutoff sample $( \mathbf{x}_{\rm cut}, \mathbf{y}_{\rm cut} )$ by following the same cutoff strategy used in Token Cutoff. For each sentence pair, the training objective of SimCut is defined as:
\begin{equation}\label{simcut}
\mathcal{L}_{simcut}(\theta) = \mathcal{L}_{ce}(\theta) + \alpha \mathcal{L}_{simkl}(\theta),
\end{equation}
where
\begin{equation*}
\mathcal{L}_{simkl}(\theta) = \text{KL}(f(\mathbf{x}, \mathbf{y}; \theta) \| f(\mathbf{x}_{\rm cut}, \mathbf{y}_{\rm cut}; \theta)).
\end{equation*}
There are only two hyper-parameters $\alpha$ and $p_{\rm cut}$ in SimCut, which greatly simplifies the hyper-parameter searching step in Token Cutoff. Note that VAT only allows the gradient to be backpropagated through the right-hand side of the KL divergence term, while the gradient is designed to be backpropagated through both sides of the KL regularization in SimCut. We can see that the constraints introduced by $\mathcal{L}_{cut}(\theta)$ and $\mathcal{L}_{kl}(\theta)$ in \eqref{token_cutoff} still implicitly hold in \eqref{simcut}:
\begin{itemize}
\item $\mathcal{L}_{cut}(\theta)$ in Token Cutoff is designed to guarantee that the output of the cutoff sample should close to the ground-truth to some extent. In SimCut, $\mathcal{L}_{ce}(\theta)$ requires the outputs of the original sample close to the ground-truth, and $\mathcal{L}_{simkl}(\theta)$ requires the output distributions of the cutoff sample close to that of the original sample. The constraint introduced by $\mathcal{L}_{cut}(\theta)$ then implicitly holds.
\item $\mathcal{L}_{kl}(\theta)$ in Token Cutoff is designed to guarantee that the output distributions of the original sample and $N$ different cutoff samples should be consistent with each other. In SimCut, $\mathcal{L}_{simkl}(\theta)$ guarantees the consistency between the output distributions of the original and cutoff samples. Even though SimCut only generates one cutoff sample at each time, different cutoff samples of the same sentence pair will be considered in different training epochs. Such constraint raised by $\mathcal{L}_{kl}(\theta)$ still implicitly holds.
\end{itemize}

\subsection{Analysis on SimCut}

\subsubsection{How Does the Simplification Affect Performance?}

\begin{table}
\centering
\begin{tabular}{c|c|c}
\hline
Method & \texttt{en}$\rightarrow$\texttt{de} & \texttt{de}$\rightarrow$\texttt{en} \\
\hline\hline
Transformer & 28.70 & 34.99 \\
VAT & 29.45 & 35.52 \\
R-Drop & 30.73 & 37.30 \\
Token Cutoff & 30.89 & 37.61 \\
\hline
SimCut & \bf 30.98 & \bf 37.81 \\
\end{tabular}
\caption{SimCut achieves the superior or comparable performance on IWSLT14 $\texttt{en}\leftrightarrow\texttt{de}$ translation tasks over the strong baselines such as VAT, R-Drop, and Token Cutoff. 
\label{simplification}}
\end{table}

We here investigate whether our simplification on Token Cutoff hurts its performance on machine translation tasks. We compare SimCut with Token Cutoff, VAT, and R-Drop \cite{liang2021r}, a strong regularization baseline that forces the output distributions of different sub-models generated by dropout to be consistent with each other. 
Table \ref{simplification} shows that SimCut achieves superior or comparable performance over VAT, R-Drop, and Token Cutoff, which clearly shows the effectiveness of our method. 
To further compare SimCut with other strong baselines in terms of training cost, we summarize the validation BLEU score along the training time on IWSLT14 \texttt{de}$\rightarrow$\texttt{en} translation task in Table \ref{training_time}. 
From the table, we can see that the BLEU score of SimCut continuously increases in the first $1500$ minutes. The results on VAT are consistent with the previous studies on adversarial overfitting, i.e., virtual adversarial training easily suffering from overfitting \cite{rice2020overfitting}. Though SimCut needs more training time to converge, the final NMT model is much better than the baseline. For the detailed training cost for each epoch, Token Cutoff costs about 148 seconds per epoch, while SimCut costs about 128 seconds per epoch. Note that the training cost of Token Cutoff is greatly influenced by the hyper-parameter $N$. We set $N$ to be $1$ in our experiments. With the increasing of $N$, the training time of Token Cutoff will be much longer. Due to the tedious and time-consuming hyper-parameter searching in Token Cutoff, we will not include its results in the following sections and show the results of SimCut directly.


\begin{table*}
\centering
\begin{tabular}{c|c|c|c|c|c|c|c|c|c|c}
\hline
Minutes & 10 & 30 & 60 & 90 & 150 & 300 & 600 & 900 & 1200 & 1500 \\
\hline\hline
Transformer & 11.51 & 31.20 & 34.19 & 34.88 & \bf 35.17 & 34.86 & 34.43 & 34.28 & 34.23 & 33.95 \\
VAT & 1.87 & 20.08 & 31.69 & 33.95 & 35.41 & 35.78 & \bf 35.81 & 35.63 & 35.17 & 34.99 \\
R-Drop & 2.11 & 26.32 & 32.81 & 34.25 & 35.88 & 36.91 & 37.18 & 37.43 & \bf 37.52 & 37.43 \\
Token Cutoff & 2.16 & 28.88 & 32.82 & 34.61 & 35.90 & 36.84 & 37.70 & 37.81 & \bf 37.93 & 37.83 \\
\hline
SimCut & 1.99 & 25.12 & 32.21 & 33.66 & 34.93 & 36.37 & 37.31 & 37.62 & 37.89 & \bf 38.10 \\
\end{tabular}
\caption{On the IWSLT14 \texttt{de}$\rightarrow$\texttt{en} validation set, the BLEU score increases over time in model training using SimCut. In contrast, the BLEU scores of the other strong baselines all stop increasing before $1500$ minutes. The results suggest that the use of SimCut can effectively alleviate the model training from overfitting. \label{training_time}}
\end{table*}

\subsubsection{How Does the Bidirectional Backpropagation Affect Performance?}

Even though the problem formulation of SimCut is similar to that of VAT, one key difference is that the gradients are allowed to be backpropagated bidirectionally in the KL regularization in SimCut. We here investigate the impact of the bidirectional backpropagation in the regularization term on the NMT performance. Table \ref{bi-back-prop} shows the translation results of VAT and SimCut with or without bidirectional backpropagation. We can see that both VAT and SimCut benefit from the bidirectional gradient backpropagation in the KL regularization.

\begin{table}
\centering
\begin{tabular}{c|c|c}
\hline
Method & \texttt{en}$\rightarrow$\texttt{de} & \texttt{de}$\rightarrow$\texttt{en} \\
\hline\hline
VAT & 29.45 & 35.52 \\
+ Bi-backpropagation & 29.69 & 36.26 \\
\hline
SimCut & 30.98 & 37.81 \\
- Bi-backpropagation & 30.29 & 36.91 \\
\end{tabular}
\caption{Bidirectional backpropagation achieves better performance on IWSLT14 $\texttt{en}\leftrightarrow\texttt{de}$ translation tasks compared with unidirectional backpropagation in the KL regularization. \label{bi-back-prop}}
\end{table}

\subsubsection{Performance on Perturbed Inputs}

Given the similar problem formulations of VAT and SimCut, it is natural to regard cutoff operation as a special perturbation and consider SimCut as a perturbation-based method. We here investigate the robustness of NMT models on the perturbed inputs. As discussed in \citet{takase2021rethinking}, simple techniques such as word replacement and word drop can achieve comparable performance to sophisticated perturbations. We hence include them as baselines to show the effectiveness of our method as follows:

\begin{itemize}[leftmargin=*]

\item {\bf UniRep}: Word replacement approach constructs a new sequence whose tokens are randomly replaced with sampled tokens. For each token in the source sentence $\mathbf{x}$, we sample $\hat{x}_i$ uniformly from the source vocabulary, and use it for the new sequence $\mathbf{x}'$ with probability $1-p'$:
\begin{equation}
x'_{i} =
\begin{cases} 
x_i,  & \mbox{with probability }p', \\
\hat{x}_i, & \mbox{with probability }1 - p'.
\end{cases}
\end{equation}
We construct $\mathbf{y}'$ from the target sentence $\mathbf{y}$ in the same manner. Following the curriculum learning strategy used in \citet{bengio2015scheduled}, we adjust $p'$ with the inverse sigmoid decay:
\begin{equation}
p'_t = \max(q, \frac{k}{k+\exp{(\frac{t}{k})}}),
\end{equation}
where $q$ and $k$ are hyper-parameters. $p'_t$ decreases to $q$ from 1, depending on the training epoch number $t$. We use $p'_t$ as $p'$ in epoch $t$. We set $q$ and $k$ to be $0.9$ and $25$ respectively in the experiments.

\item {\bf WordDrop}: Word drop randomly applies the zero vector instead of the word embedding $e(x_i)$ or $e(y_i)$ for the input token $x_i$ or $y_i$ \cite{gal2016theoretically}. For each token in both source and target sentences, we keep the original embedding with the probability $\beta$ and set it to be the zero vector otherwise. We set $\beta$ to be $0.9$ in the experiments.


\end{itemize}

\begin{table}
\centering
\begin{tabular}{c|c|c|c|c}
\hline
Method & \multicolumn{4}{|c}{probability} \\
 & 0.00 & 0.01 & 0.05 & 0.10 \\ 
\hline\hline
Transformer & 34.99 & 34.01 & 30.38 & 25.70 \\ 
UniRep & 35.67 & 34.91 & 31.54 & 27.24 \\ 
WordDrop & 35.65 & 34.73 & 31.22 & 26.46 \\ 
VAT & 35.52 & 34.65 & 30.48 & 25.44 \\ 
R-Drop & 37.30 & 36.24 & 32.27 & 27.19 \\ 
\hline
SimCut & \bf 37.81 & \bf 36.94 & \bf 33.16 & \bf 27.93 \\
\end{tabular}
\caption{\label{noisy}
The model trained by SimCut achieves high robustness on the perturbed test set and high performance on the clean test set. Entries represent BLEU scores on IWSLT14 \texttt{de}$\rightarrow$\texttt{en} test set when we inject perturbations to source sentences with different probability.
}
\end{table}

We construct noisy inputs by randomly replacing words in the source sentences based on a pre-defined probability. If the probability is $0.0$, we use the original source sentence. If the probability is $1.0$, we use completely different sentences as source sentences. We set the probability to be $0.00$, $0.01$, $0.05$, and $0.10$ in our experiments. We randomly replace each word in the source sentence with a word uniformly sampled from the vocabulary. We apply this procedure to IWSLT14 \texttt{de}$\rightarrow$\texttt{en} test set. 
Table~\ref{noisy} shows the BLEU scores of each method on the perturbed test set. Note that the BLEU scores are calculated against the original reference sentences. We can see that all methods improve the robustness of the NMT model, and SimCut achieves the best performance among all the methods on both the clean and perturbed test sets.  
The performance results indicate that SimCut could be considered as a strong baseline for the perturbation-based method for the NMT model.

\begin{table*}
\centering
\begin{tabular}{c|l}
\hline
\multirow{2}{*}{Input} & wir denken ({\color{red}festgelegten}), dass wir in der realität nicht so gut \\
 & sind wie in spielen. \\
\hline
Reference & we feel that we are not as good in reality as we are in games. \\
\hline
\hline
\citet{vaswani2017attention} on Input & we think we're not as good in reality as we are in games. \\
\hline
on Noisy Input & we realized that we weren't as good as we were in real life. \\
\hline
\hline
SimCut on Input & we think in reality, we're not as good as we do in games. \\
\hline
on Noisy Input & we realized that we're not as good in reality as we are in games.\\
\hline
\end{tabular}
\caption{\label{noisy-example}
SimCut is more robust to small perturbations in an authentic context. SimCut captures the translation of ``in spielen'' under the noisy input while the vanilla Transformer ignores the translation of ``in spielen'' due to the replacement of ``denken'' with ``festgelegten''.
}
\end{table*}

As shown in Table~\ref{noisy-example}, the baseline model completely ignores the translation of ``in spielen (in games)'' due to the replacement of ``denken (think)'' with ``festgelegten (determined)'' in the source sentence. In contrast, our model successfully captures the translation of ``in spielen'' under the noisy input. This result shows that our model is more robust to small perturbations in an authentic context.

\subsubsection{Effects of $\alpha$ and $p_{\rm cut}$}

\begin{figure}[h]
\centering
\includegraphics[scale=0.64]{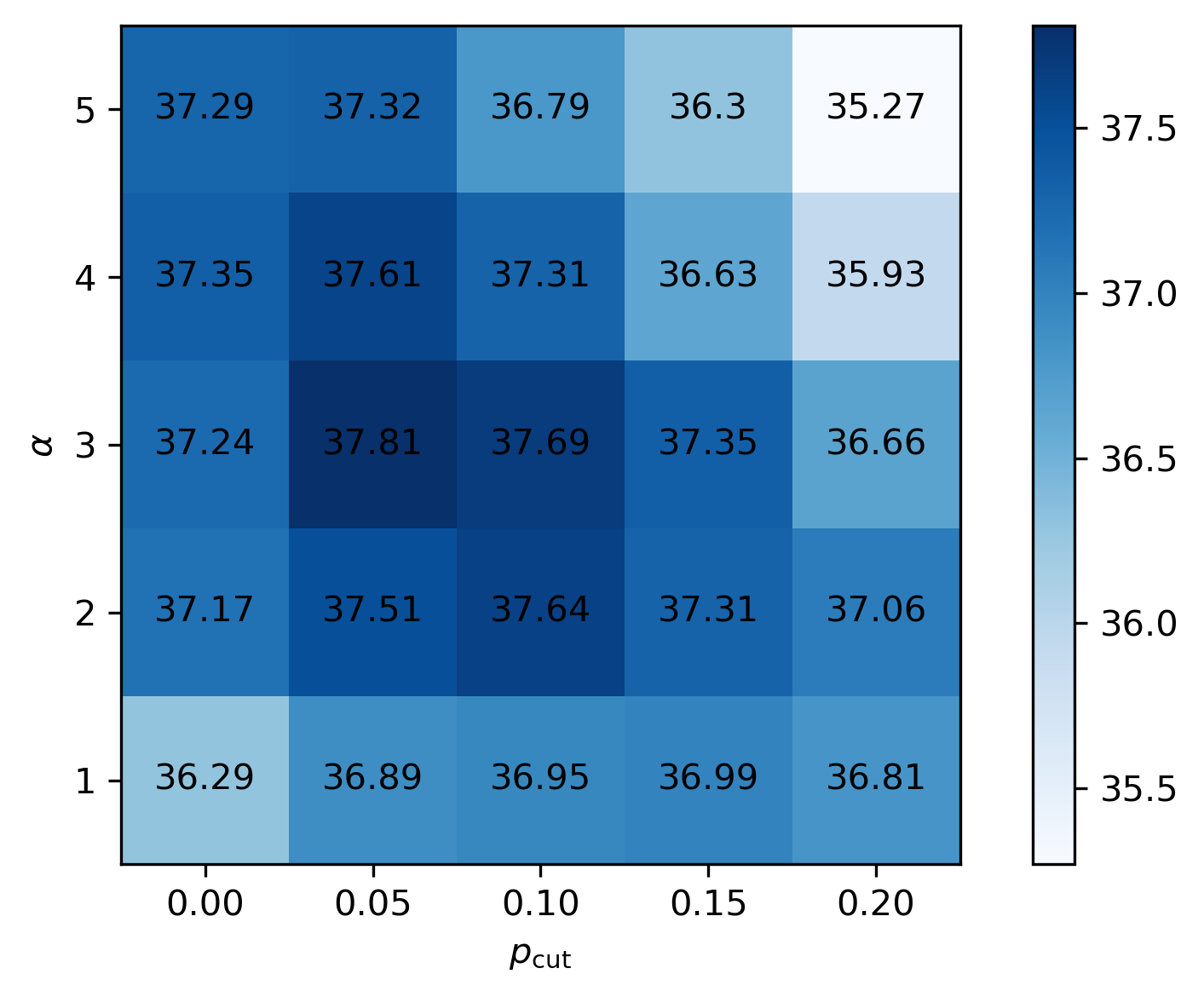}
\caption{BLEU scores with different $\alpha$ and $p_{\rm cut}$ on IWSLT14 \texttt{de}$\rightarrow$\texttt{en} dataset.}
\label{fig: alpha_p}
\end{figure}

We here investigate the impact of the scalar hyper-parameters $\alpha$ and $p_{\rm cut}$ in SimCut. $\alpha$ is a penalty parameter that controls the regularization strength in our optimization problem. $p_{\rm cut}$ controls the percentage of the cutoff perturbations in SimCut. We here vary $\alpha$ and $p_{\rm cut}$ in $\{1, 2, 3, 4, 5\}$ and $\{0.00, 0.05, 0.10, 0.15, 0.20\}$ respectively and conduct the experiments on the IWSLT14 \texttt{de}$\rightarrow$\texttt{en} dataset. Note that SimCut is simplified to R-Drop approximately when $p_{\rm cut}=0.00$. The test BLEU scores are reported in Figure \ref{fig: alpha_p}. By checking model performance under different combinations of $\alpha$ and $p_{\rm cut}$, we have the following observations: 1) A too small $\alpha$ (e.g., $1$) cannot achieve as good performance as larger $\alpha$ (e.g., $3$), indicating a certain degree of regularization strength during NMT model training is conducive to generalization. Meanwhile, an overwhelming regularization ($\alpha=5$) is not plausible for learning NMT models. 2) When $\alpha=3$, the best performance is achieved when $p_{\rm cut}=0.05$, and $p_{\rm cut}=0.00$ performs sub-optimal among all selected probabilities. Such an observation demonstrates that the cutoff perturbation in SimCut can effectively promote the generalization compared with R-Drop. 

\subsubsection{Is SimCut Compatible with the Pretrained Language Model?}

\begin{table}
\centering
\begin{tabular}{c|c}
\hline
Method & \texttt{de}$\rightarrow$\texttt{en} \\
\hline\hline
Transformer  & 32.4 \\
mBART & 38.5 \\
\hline
mBART with SimCut & \bf 39.3 \\
\end{tabular}
\caption{SimCut achieves better performance on IWSLT14 \texttt{de}$\rightarrow$\texttt{en} translation task compared with the standard finetuning approach based on mBART. \label{mbart}}
\end{table}

The multilingual sequence-to-sequence pretrained language models \cite{song2019mass, liu2020multilingual, xue-etal-2021-mt5} have shown impressive performance on machine translation tasks, where the pretrained models generally learn the knowledge from the large-scale monolingual data. It is interesting to investigate whether SimCut can gain performance improvement based on the pretrained language model. We adopt mBART \cite{liu2020multilingual} as the backbone model, which is a sequence-to-sequence denoising auto-encoder pretrained on CC25 Corpus\footnote{https://github.com/pytorch/fairseq/tree/main/examples/ mbart}. We conduct experiments on IWSLT14 \texttt{de}$\rightarrow$\texttt{en} dataset and only remove the duplicated sentence pairs following mBART50 \cite{tang2021multilingual} in the data preprocessing step. The source and target sentences are jointly tokenized into sub-word units with the 250K SentencePiece \cite{kudo-richardson-2018-sentencepiece} vocabulary of mBART. We use case-sensitive sacreBLEU \cite{post-2018-call} to evaluate the translation quality, and the methods applied in the experiments are as follows:
\begin{itemize}
\item Transformer: The Transformer model is randomly initialized and trained from scratch. We utilize the same model and training configurations discussed in Section \ref{data_baseline}.
\item mBART: The Transformer model is directly finetuned from mBART. We utilize the default training configurations of mBART.
\item mBART with SimCut: The Transformer model is finetuned from mBART with SimCut regularization. We utilize the default training configurations of mBART.
\end{itemize}
From Table \ref{mbart} we can see that SimCut could further improve the translation performance of mBART, which again shows the effectiveness and universality of our method.


\subsection{Training Strategy: Bidirectional Pretrain and Unidirectional Finetune}

Bidirectional pretraining is shown to be very effective to improve the translation performance of the unidirectional NMT system \cite{ding2021improving, xu-etal-2021-bert}. The main idea is to pretrain a bidirectional NMT model at first and use it as the initialization to finetune a unidirectional NMT model. Assume we want to train an NMT model for ``English$\rightarrow$German'', we first reconstruct the training sentence pairs to ``English$+$German$\rightarrow$German$+$English'', where the training dataset is doubled. We then firstly train a bidirectional NMT model with the new training sentence pairs:
\begin{equation}
\mathop{\mathbb{E}}\limits_{(\mathbf{x}, \mathbf{y}) \in \mathcal{S}} [\ell(f(\mathbf{x}, \mathbf{y}; \theta), \ddot{\mathbf{y}}) + \ell(f(\mathbf{y}, \mathbf{x}; \theta), \ddot{\mathbf{x}})],
\end{equation}
and finetune the model with ``English$\rightarrow$German'' direction. We follow the same training strategy in \citet{ding2021improving} and apply SimCut regularization to both pretraining and finetuning procedures. Table \ref{bit} shows that bidirectional pretraining and unidirectional finetuning strategy with SimCut regularization could achieve superior performance compared with strong baseline such as R-Drop.  

\begin{table}
\centering
\begin{tabular}{c|c|c}
\hline
Method & \texttt{en}$\rightarrow$\texttt{de} & \texttt{de}$\rightarrow$\texttt{en} \\
\hline\hline
Transformer & 28.70 & 34.99 \\
\hline
Bi-Pretrain & 28.94 & 35.64 \\
+ Finetune & 28.82 & 35.66 \\
\hline
Bi-R-Drop Pretrain & 30.30 & 37.01 \\
+ R-Drop Finetune & 30.85 & 37.55 \\
\hline
Bi-SimCut Pretrain & 30.57 & 37.70 \\
+ SimCut Finetune & \bf 31.16 & \bf 38.37 \\
\end{tabular}
\caption{Bidirectional pretrain and unidirectional finetune results on IWSLT14 $\texttt{en}\leftrightarrow\texttt{de}$ datasets. Note that the results of bidirectional pretrain are from one model for dual-directional translations. \label{bit}}
\end{table}

\paragraph{Comparison with Existing Methods}

We summarize the recent results of several existing works on IWSLT14 \texttt{en}$\leftrightarrow$\texttt{de} benchmark in Table \ref{benchmark}. 
The existing methods vary from different aspects, including Virtual Adversarial Training \cite{sano2019effective}, Mixed Tokenization for NMT \cite{wu2020sequence}, Unified Dropout for the Transformer model \cite{wu2021unidrop}, Regularized Dropout \cite{liang2021r}, and BiBERT \cite{xu-etal-2021-bert}. We can see that our approach achieves an improvement of $2.92$ BLEU score over \citet{vaswani2017attention} and surpass the current SOTA method BiBERT that incorporates large-scale pretrained model, stochastic layer selection, and bidirectional pretraining. Given the simplicity of Bi-SimCut, we believe it could be considered as a strong baseline for the NMT task.

\begin{table}
\centering
\begin{tabular}{c|c|c|c}
\hline
Method & \texttt{en}$\rightarrow$\texttt{de} & \texttt{de}$\rightarrow$\texttt{en} & Average \\
\hline\hline
Transformer & 28.70 & 34.99 & 31.85 \\
VAT & 29.45 & 35.52 & 32.49 \\
Mixed Rep$^{\dagger}$ & 29.93 & 36.41 & 33.17 \\
UniDrop$^{\dagger}$ & 29.99 & 36.88 & 33.44\\
R-Drop & 30.73 & 37.30 & 34.02 \\
BiBERT$^{\dagger}$ & 30.45 & \bf 38.61 & 34.53 \\
\hline
Bi-SimCut & \bf 31.16 & 38.37 & \bf 34.77 \\
\end{tabular}
\caption{Our method achieves the superior performance over the existing methods on the IWSLT14 \texttt{en}$\leftrightarrow$\texttt{de} translation benchmark. $\dagger$ denotes the numbers are reported from the corresponding papers, others are based on our runs. \label{benchmark}}
\end{table}

\section{Standard Resource Scenario}\label{standard-resource}

We here investigate the performance of Bi-SimCut on the larger translation benchmark compared with the IWSLT14 benchmark.

\begin{table*}
\centering
\begin{tabular}{c|c|c|c}
\hline
Method & \texttt{en}$\rightarrow$\texttt{de} & \texttt{de}$\rightarrow$\texttt{en} & Average \\
\hline\hline
Transformer + Large Batch$^{\dagger}$ \cite{ott2018scaling} & 29.30 & - & - \\
Evolved Transformer$^{\dagger}$ \cite{so2019evolved} & 29.80 & - & - \\
BERT Initialization (12 layers)$^{\dagger}$ \cite{rothe2020leveraging} & 30.60 & 33.60 & 32.10 \\
BERT-Fuse$^{\dagger}$ \cite{zhu2020incorporating} & 30.75 & - & - \\
R-Drop \cite{liang2021r} & 30.13 & 34.54 & 32.34 \\
BiBERT$^{\dagger}$ \cite{xu-etal-2021-bert} & \bf 31.26 & 34.94 & \bf 33.10 \\
\hline
SimCut & 30.56 & 34.86 & 32.71 \\
Bi-SimCut Pretrain & 30.10 & 34.42 & 32.26 \\
+ SimCut Finetune & 30.78 & \bf 35.15 & 32.97 \\
\end{tabular}
\caption{Our method achieves the superior or comparable performance over the existing methods on the WMT14 \texttt{en}$\leftrightarrow$\texttt{de} translation benchmark. $\dagger$ denotes the numbers are reported from \citet{xu-etal-2021-bert}, others are based on our runs. \label{ende}}
\end{table*}

\subsection{Dataset Description and Model Configuration}

For the standard resource scenario, we evaluate NMT models on the WMT14 English-German dataset, which contains 4.5M parallel sentence pairs. We combine newstest2012 and newstest2013 as the validation set and use newstest2014 as the test set. We collect the pre-processed data from \citet{xu-etal-2021-bert}'s release\footnote{https://github.com/fe1ixxu/BiBERT}, where a shared dictionary with 52K BPE types is built. We apply a standard Transformer Big model with 6 encoder and decoder layers, 16 attention heads, embedding size 1024, and FFN layer dimension 4096. We apply cross-entropy loss with label smoothing rate $0.1$ and set max tokens per batch to be $4096$. We use Adam optimizer with Beta $(0.9, 0.98)$, $4000$ warmup updates, and inverse square root learning rate scheduler with initial learning rates $1e^{-3}$. We decrease the learning rate to $5e^{-4}$ in the finetuning stage. We select the dropout rate from $0.3$, $0.2$, and $0.1$ based on the validation performance. We use beam search decoding with beam size $4$ and length penalty $0.6$. We train all models until convergence on 8 NVIDIA Tesla V100 GPUs. All reported BLEU scores are from a single model. 

\subsection{Results}

We report test BLEU scores of all comparison methods and our approach on the WMT14 dataset in Table \ref{ende}. With Bi-SimCut bidirectional pretraining and unidirectional finetuning procedures, our NMT model achieves strong or SOTA BLEU scores on \texttt{en}$\rightarrow$\texttt{de} and \texttt{de}$\rightarrow$\texttt{en} translation benchmarks. During the NMT training process, we fix $p_{\rm cut}$ to be $0.05$ and tune the hyper-parameter $\alpha$ in both R-Drop and SimCut based on the performance on the validation set. Note that the BLEU scores of R-Drop are lower than that reported in \citet{liang2021r}. Such gap might be due to the different prepossessing steps used in \citet{liang2021r} and \citet{xu-etal-2021-bert}. It is worth mentioning that Bi-SimCut outperforms BiBERT on \texttt{de}$\rightarrow$\texttt{en} direction even though BiBERT incorporates bidirectional pretraining, large-scale pretrained contextualized embeddings, and stochastic layer selection mechanism.

\section{High Resource Scenario}\label{high-resource}

To investigate the performance of Bi-SimCut on the distant language pairs which naturally do not share dictionaries, we here discuss the effectiveness of Bi-SimCut on the Chinese-English translation task.

\subsection{Dataset Description and Model Configuration}

For the high resource scenario, we evaluate NMT models on the WMT17 Chinese-English dataset, which consists of 20.2M training sentence pairs, and we use newsdev2017 as the validation set and newstest2017 as the test set. We firstly build the source and target vocabularies with 32K BPE types separately and treat them as separated or joined dictionaries in our experiments. We apply the same Transformer Big model and training configurations used in the WMT14 experiments. We use beam search decoding with beam size $5$ and length penalty $1$. We train all models until convergence on 8 NVIDIA Tesla V100 GPUs. All reported BLEU scores are from a single model. 

\subsection{Results}

\begin{table}
\centering
\begin{tabular}{c|c|c}
\hline
Method & $\texttt{share}$ & \texttt{zh}$\rightarrow$\texttt{en} \\
\hline\hline
Transformer & $\text{\sffamily x}$ & 25.53 \\
Transformer & $\checkmark$ & 25.31 \\
\hline
SimCut & $\text{\sffamily x}$ & 26.86 \\ 
SimCut & $\checkmark$ & 26.74 \\ 
Bi-SimCut Pretrain & $\checkmark$ & 26.13 \\
+ SimCut Finetune & $\checkmark$ & \bf 27.17 \\
\end{tabular}
\caption{Our method achieves strong performance on the WMT17 \texttt{zh}$\rightarrow$\texttt{en} translation benchmark. $\texttt{share}$ denotes whether a shared dictionary is applied. \label{zhen}}
\end{table}

We report test BLEU scores of the baselines and our approach on the WMT17 dataset in Table \ref{zhen}. Note that $\texttt{share}$ means the embedding matrices for encoder input, decoder input and decoder output are all shared.
The NMT models with separated dictionaries perform slightly better than those with the shared dictionary. We can see that our approach significantly improves the translation performance. In particular, Bi-SimCut achieves more than $1.6$ BLEU score improvement over \citet{vaswani2017attention}, showing the effectiveness and universality of our approach on the distant language pair in the NMT task.

\section{Related Work}

\paragraph{Adversarial Perturbation}

It is well known that neural networks are sensitive to noisy inputs, and adversarial perturbations are firstly discussed in the filed of image processing \cite{szegedy2013intriguing, goodfellow2014explaining}. SimCut could be regarded as a perturbation-based method for the robustness research. 
In the field of natural language processing, \citet{miyato2016adversarial} consider adversarial perturbations in the embedding space and show its effectiveness on the text classification tasks. For the NMT tasks, \citet{sano2019effective} and \citet{wang2019improving} apply adversarial perturbations in the embedding space during training of the encoder-decoder NMT model. \citet{cheng2019robust} leverage adversarial perturbations and generate adversarial examples by replacing words in both source and target sentences. They introduce two additional language models for both sides and a candidate word selection mechanism for replacing words in the sentence pairs. \citet{takase2021rethinking} compare perturbations for the NMT model in view of computational time and show that simple perturbations are sufficiently effective compared with complicated adversarial perturbations.

\paragraph{Consistency Training}

Besides perturbation-based methods, our approach also highly relates to a few works of model-level and data-level consistency training in the NMT field. Among them, the most representative methods are R-Drop \cite{liang2021r} and Cutoff \cite{shen2020simple}. R-Drop studies the intrinsic randomness in the NMT model and regularizes the NMT model by utilizing the output consistency between two dropout sub-models with the same inputs. Cutoff considers consistency training from a data perspective by regularizing the inconsistency between the original sentence pair and the augmented samples with part of the information within the input sentence pair being dropped. Note that Cutoff takes the dropout sub-models into account during the training procedure as well. We want to emphasize that SimCut is not a new method, but a version of Cutoff simplified and adapted for the NMT tasks.

\section{Conclusion}

In this paper, we propose Bi-SimCut: a simple but effective two-stage training strategy to improve NMT performance. Bi-SimCut consists of bidirectional pretraining and unidirectional finetuning procedures equipped with SimCut regularization for improving the generality of the NMT model. Experiments on low (IWSLT14 \texttt{en}$\leftrightarrow$\texttt{de}), standard (WMT14 \texttt{en}$\leftrightarrow$\texttt{de}), and high (WMT17 \texttt{zh}$\rightarrow$\texttt{en}) resource translation benchmarks demonstrate Bi-SimCut and SimCut's capabilities to improve translation performance and robustness. Given the universality and simplicity of Bi-SimCut and SimCut, we believe: 1) SimCut could be regarded as a perturbation-based method, and it could be used as a strong baseline for the robustness research. 2) Bi-SimCut outperforms many complicated methods which incorporate large-scaled pretrained models or sophisticated mechanisms, and it could be used as a strong baseline for future NMT research. We hope researchers of perturbations and NMT could use SimCut and Bi-SimCut as strong baselines to make the usefulness and effectiveness of their proposed methods clear. For future work, we will explore the effectiveness of SimCut and Bi-SimCut on more sequence learning tasks, such as multilingual machine translation, domain adaptation, text classification, natural language understanding, etc.

\section*{Acknowledgements}

We would like to thank the anonymous reviewers for their insightful comments.

\bibliography{anthology,custom}
\bibliographystyle{acl_natbib}


\end{document}